\documentclass{article}

\PassOptionsToPackage{numbers, compress, sort}{natbib}

\usepackage[dandb, final]{neurips_2025}

\usepackage[utf8]{inputenc} %
\usepackage[T1]{fontenc}    %
\usepackage{hyperref}       %
\usepackage{url}            %
\usepackage{booktabs}       %
\usepackage{amsfonts}       %
\usepackage{nicefrac}       %
\usepackage{microtype}      %
\usepackage[dvipsnames]{xcolor}         %

\usepackage{amsmath}
\usepackage{amssymb}
\usepackage{bm}
\usepackage{bbm}
\usepackage{xspace}
\usepackage{graphicx}
\usepackage{comment}
\usepackage[capitalize]{cleveref}

\usepackage{subcaption}
\definecolor{myred}{RGB}{220,0,0} \definecolor{mydarkgreen}{RGB}{0,100,0}

\DeclareMathOperator*{\argmax}{arg\,max}

\newcommand*{\ie}{i.e.,\@\xspace}
\newcommand*{\eg}{e.g.,\@\xspace}

\newcommand{\datasetname}{CleverBirds\@\xspace}

\newcommand{\numusers}{40,144}

\newcommand{\uniquespeciesnum}{10,779}
\newcommand{\uniquespeciesnumIcnlChoices}{11,142}

\newcommand{\totalRowsDataset}{17,859,392}

\title{\datasetname:  A Multiple-Choice Benchmark for Fine-grained Human Knowledge Tracing}

\author{%
  Leonie Bossemeyer\textsuperscript{1} \quad
  Samuel Heinrich\textsuperscript{2} \quad
  Grant Van Horn\textsuperscript{3} \quad
  Oisin Mac Aodha\textsuperscript{1} \\
  \textsuperscript{1}University of Edinburgh \quad \quad
  \textsuperscript{2}Cornell University \quad \quad
  \textsuperscript{3}UMass Amherst   
}

\begin{document}

\maketitle

\begin{abstract}
Mastering fine-grained visual recognition, essential in many expert domains, can require that specialists undergo years of dedicated training. Modeling the progression of such expertize in humans remains challenging, and accurately inferring a human learner’s knowledge state is a key step toward understanding visual learning. We introduce \datasetname{}, a large-scale knowledge tracing benchmark for fine-grained bird species recognition. Collected by the citizen-science platform eBird, it offers insight into how individuals acquire expertize in complex fine-grained classification. More than 40,000 participants have engaged in the quiz, answering over 17 million multiple-choice questions spanning over 10,000 bird species, with long-range learning patterns across an average of 400 questions per participant. We release this dataset to support the development and evaluation of new methods for visual knowledge tracing. We show that tracking learners' knowledge is challenging, especially across participant subgroups and question types, with different forms of contextual information offering varying degrees of predictive benefit. \datasetname{} is among the largest benchmark of its kind, offering a substantially higher number of learnable concepts. With it, we hope to enable new avenues for studying the development of visual expertize over time and across individuals.
\end{abstract}

\section{Introduction}
\vspace{-5pt}

Dedicated practice and expert instruction from knowledgeable teachers are essential ingredients for students tasked with mastering new subjects and concepts.   
However, providing access to high quality, yet affordable, instruction at scale is time consuming and expensive~\cite{bloom19842}. 
As a result, researchers have looked towards alternative, computer-assisted~\cite{psotka1988intelligent}, tools in an attempt to overcome these hurdles.

At the heart of an effective automated tutoring system is a computational model of the human learner. 
The goal of these models is to observe the learner as they engage with the teaching material at hand, represent the learners' knowledge state, and estimate any potential knowledge gaps they may have.
This task, also known as knowledge tracing (KT), has a long history in the literature~\cite{abdelrahman2023knowledge}. 
Early solutions modeled human mastery of the material being learned via latent variable probabilistic models~\cite{Cen2008, Cen2007Jun}. 
More recent approaches have advocated for the use of deep learning-based solutions \cite{piech2015deep} which, while effective at capturing more complex relationships, can require large quantities of data to train. 
However, current datasets for quantifying the performance of different KT methods are typically concentrated around a small number of subjects such as mathematics~\cite{feng2009addressing, chang2015modeling, liu2023xes3g5m, wang2020diagnostic}, programming~\cite{pandey2020rkt}, and language learning~\cite{kim2024kt, choi2020ednet}. 

In this work, we attempt to address a gap in the existing available benchmark datasets for KT. 
This is motivated by the fact that there are a large number of domains where learners wish to learn visual identification skills, \eg in medicine, art, and biology, to name a few. 
Many of the tasks in these domains can be posed as classification problems, where the human learner attempts to learn the decision boundaries between different concepts (\ie different semantic classes). 
One such domain is animal species classification. 
For example, there are now a number of online platforms where members of the public report sightings and locations of different species from all around the world, which in turn is providing valuable data for science~\cite{callaghan2021three}. 
The challenge for the participants in these projects is that the number of visual concepts (\ie species) can be very large. 
For example, even if only restricted to the case of birds, there are over 11,000 different species worldwide. 
Compounding this difficulty is the fact that discriminating between certain species can require very fine-grained knowledge~\cite{wei2021fine} as some species can look very similar to others.

\begin{figure}[t]
\centering
\includegraphics[width=1.0\textwidth]{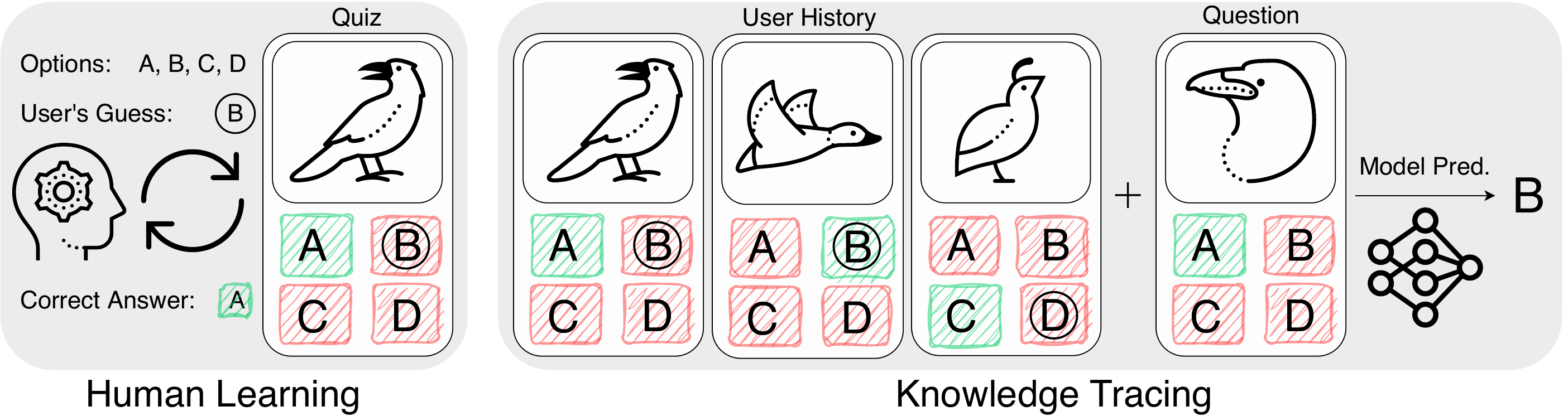}
\caption{{\bf (Left) Human Learning}. Participants learn from the quiz questions contained in \datasetname{} through repeated interactions. 
For each question, participants are presented with an image of a bird species and a list of possible  species  names (here \{`A', `B', `C', `D'\}), which may include the correct answer. After making a guess, they receive feedback in the form of the correct answer (here `A'). 
This process is repeated for multiple questions.
{\bf (Right) Knowledge Tracing}. We illustrate the prediction task, in which a model is given a participant’s interaction history together with the current question’s image, options, and correct answer, and is tasked with predicting the participant’s guess.}
\label{fig:task_overview}
\end{figure}

\begin{figure}[t]
\centering
\includegraphics[width=1.0\textwidth]{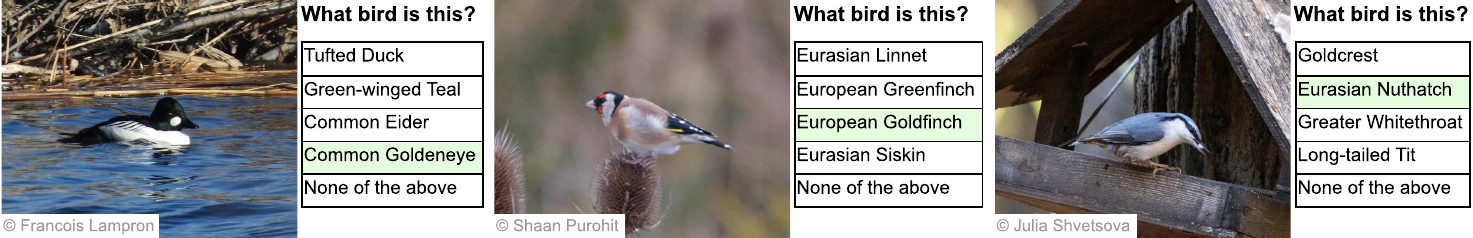}
\caption{Three examples of the types of quiz questions found in our \datasetname dataset. 
In each case, there are four options representing different species and an additional ``None of the above option''. 
The correct answer is indicated in
\textcolor{YellowGreen}{green}. 
Any of five options are valid answers and the set of candidate species provided in the option set are different for each question. 
}
\label{fig:quiz_examples}
\end{figure}

To advance the development of KT methods in the context of fine-grained classification tasks we introduce the \datasetname{} dataset. 
\datasetname{} embodies a challenging real-world classification task and contains a large number of interactions generated by human participants who are attempting to learn how to identify different bird species from images. 
The core task is depicted in \cref{fig:task_overview} where a learner is presented with a sequence of multiple-choice questions and the aim is for them to correctly identify the bird species depicted in the images shown. 
Example questions from the dataset are shown in \cref{fig:quiz_examples}.

We make the following contributions: 
(1) We introduce \datasetname, a new large-scale benchmark for visual knowledge tracing. Our dataset contains over 10,000 visual unique  concepts and more than 17 million total interactions from over 40,000 unique participants, with  half of the participants having answered over 100 questions each. It provides a new benchmark for obtaining insights into human learning in the context of fine-grained visual classification. 
(2) We quantitatively evaluate a range of computational approaches on \datasetname and demonstrate that it is a challenging benchmark, not only to human participants, but also to the computational methods tested. We evaluate these baseline methods under varying levels of input context, and show how different types of information can impact predictive accuracy. 
Links to the dataset and code are available at \url{https://cleverbirds-benchmark.github.io}.

\section{Related Work}
\vspace{-5pt}
{\bf Knowledge Tracing (KT).}  
KT methods aim to model student knowledge acquisition over time such that they can predict how a given student will perform on future interactions~\cite{abdelrahman2023knowledge}.  
Effective models of human learning have a wide array of  applications in the context of intelligent tutoring systems~\cite{psotka1988intelligent} and machine teaching~\cite{zhu2015machine}. %
Traditional KT approaches are based on probabilistic models of student mastery~\cite{corbett1994knowledge}, but multiple extensions have been proposed to address some of the simplifying assumptions that were commonly made by earlier models, \eg  by incorporating individual-specific learnable parameters~\cite{yudelson2013individualized}, by estimating concept difficulty~\cite{pardos2011kt}, or by modeling more complex dependencies between concepts~\cite{kaser2017dynamic}.

More recently, there has been a growth in the number of deep learning-based KT approaches proposed~\cite{song2022survey}. 
Various architectures have been explored such as recurrent networks~\cite{piech2015deep,nagatani2019augmenting,yeung2018addressing}, graph neural networks~\cite{nakagawa2019graph,yang2021gikt,tong2020structure}, 
attention-based models~\cite{pandey2019self,ghosh2020context,pandey2020rkt}, 
memory augmented models~\cite{zhang2017dynamic,abdelrahman2019knowledge}, 
 hierarchical approaches~\cite{wang2019deep,liu2021hierarchical}, and explainable methods~\cite{bai2024survey}. 
There have also been attempts to utilize contextual knowledge extracted from large language models to better encode interactions between concepts and questions~\cite{lee2024language,fu2024sinkt}.   
Deep KT methods can capture complex interactions and longer range temporal dependencies, but at the cost of requiring larger training datasets~\cite{gervet2020deep}.

{\bf Datasets of Human Learning.} 
There are a large number of benchmark datasets that have been utilized to quantify the performance of different KT methods. 
These datasets are typically distinguished in terms of the number of human learners (\ie participants), the number of knowledge components/concepts (\eg `subtraction' could be a concept in the context of mathematics), the number of unique questions, and the total number of interactions (\ie each student may not attempt all questions).  
Synthetic datasets have the advantage of enabling controlled testing as the underlying generative process is known~\cite{piech2015deep}.  
However, evaluation in simulation is not a substitute for performing experiments using data from real human participants. 
The most common source of data comes from mathematical education, \eg~\cite{feng2009addressing,koedinger2010data,stamper2010kddcup2005, stamper2010kddcup2006,wang2020diagnostic}. 
The largest of these datasets consist of millions of interactions with thousands of students and can also contain additional auxiliary information~\cite{liu2023xes3g5m}. 
However, existing datasets are typically limited in the number of overall concepts contained within them. 
Beyond mathematics, other popular domains include linguistics~\cite{choi2020ednet},  programming~\cite{fu2024sinkt,kim2024kt}, and general education games~\cite{kim2024kt}.

Most relevant to our benchmark are the small number of datasets targeting image classification tasks. 
While not precisely image data, \cite{crowston2019knowledge} perform experiments on a dataset containing spectrograms derived from  gravitational wave observations from the Gravity Spy citizen science project~\cite{zevin2017gravity}. 
The goal for participants is to classify each spectrogram into one of a discrete set of classes representing 21 different types of `glitches'.  
This dataset does not explicitly target a learning setting, \ie it is not necessarily the case that the participants get better over time. 
The authors of~\cite{kondapaneni2022visual} introduce three datasets for evaluating human learning of fine-grained visual concepts. 
Their datasets contain images of five species of butterflies, three classes of conditions of human retinas, and three classes of synthetic `greebles'. 
Each dataset only contains 6,750 interactions which were obtained using participants from an online crowd working platform. 
For each of the three datasets, there are less than one thousand total images.  
We summarize the statistics of current KT datasets in \cref{tab:kt_datasets}. 

In contrast to existing image-based datasets, our \datasetname benchmark contains a much larger number of possible concepts that can be learned, \ie~{\uniquespeciesnum{}} different species of birds from all around the world. 
In addition, the participant pool spans a range of expertize levels and is sourced from engaged individuals that volunteered to participate based on their interest in the problem domain.

\vspace{-5pt}
\section{\datasetname{} Dataset}
\label{sec:dataset} 
\vspace{-5pt}

Here we describe our \datasetname{} dataset. 
We outline the original collection protocol, the steps we undertook to refine it, and we describe the high-level statistics of the dataset.  

{\bf Quiz Composition.} The \datasetname{} dataset is sourced from an online  bird species identification quiz~\cite{eBirdQuiz} created by the citizen science project eBird~\cite{sullivan2009ebird}. 
In the quiz, users (\ie participants) are shown an image and asked to guess which bird species,  from a list of options, is present in the image shown (see \cref{fig:quiz_examples} for examples). 
It was first published online in March 2018 to encourage users to provide image quality ratings for new image uploads. 
The images contained within the quiz are sourced from citizen scientists who upload images, along with a proposed species label indicating the species present in the image, to the Macaulay Library \cite{macaulaylibrary}. %
This proposed species label is cross-referenced with a list of species expected to be found in the geographic location, and a volunteer expert reviewer is consulted for unlikely cases (\ie an image identified as containing a specific species that is not typically found in that area). 
Data labeling errors from these types of citizen science efforts can occur, but are low. 
For example, on iNaturalist, the error rate for bird species identification has been found to be 3.3\%~\cite{iNaturalistAccExp}. %

Quiz images are sampled from images that have been uploaded within the last 5 to 365 days and contain proportions of both images with no quality ratings, and images that have quality ratings of at least 2.4 out of 5 possible stars. 
This allows for unrated images to be quality-labeled through use of the quiz, while keeping enough high quality images in the quiz for the users to learn from. Users are asked to rate image quality by sharpness, visibility of the bird, size of the photo and watermarks, while allowing for flocks and birds in-hand~\cite{ratequalityebird}. Additionally, users are encouraged to skip questions that are unanswerable, for example because multiple bird species are visible within one image. To further increase difficulty, the candidate answer options are selected to be taxonomically similar to the correct species. Specifically, options are drawn from a sliding window over the taxonomic list centered around the true species present in the image shown.

{\bf Graphical Interface.} Users initiate a quiz by selecting parameters such as a location, time of year, and species prevalence, which are used to generate quiz questions. 
For example, if a user selected Edinburgh, Scotland, May 15th, and `likely' species, they would only be presented with questions featuring common birds that would be expected to be found in that region at that time of year. 
Note, users have the option to select an audio quiz, whereby audio recordings are played instead of them being shown images. 
However, we only use the image quiz data as it is much more prevalent. 
Users are then guided through a set of 20 multiple-choice questions. For each question, an image of a bird is shown along with five answer choices: four species names and a ``None of the above'' option. Users can optionally skip any question. 
After submitting an answer, the correct species is revealed, and they are asked to rate the image quality on a scale of 1 to 5. 

\begin{figure}[t]
\centering
\includegraphics[width=\textwidth]{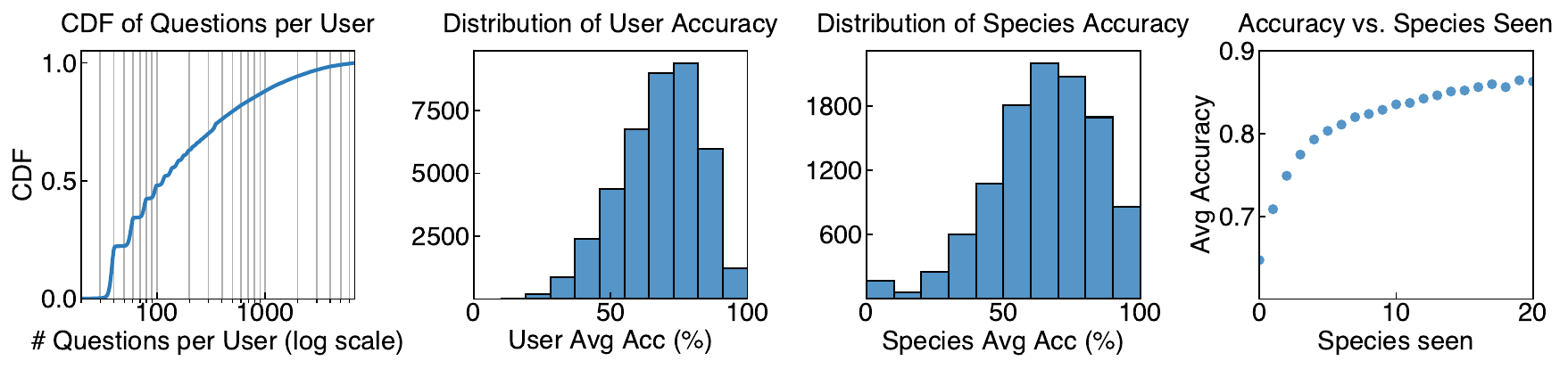}
\vspace{-10pt}
\caption{Left to right: Cumulative distribution of quizzes attempted per user on a log scale, distribution of users' average accuracies,  distribution of species-wise average user accuracies, and  average user accuracy by number of prior exposures to a species. }
\label{fig:dataset_stats}
\end{figure}

{\bf Data Filtering.} \datasetname{} is based on all quizzes completed online from March 14th 2018 to October 8th, 2024. 
To prevent overfitting on specific users, we split the dataset by user ID into training, validation, and test sets. Aiming for a 70/15/15 split, users are assigned to the training set until it comprises 70\% of interactions, then we continue to the validation set and finally the test set. This results in 28,100 users in training, 6,021 in validation, and 6,023 in test, corresponding to 70.6\%, 14.6\%, and 14.8\% of interactions, respectively. 
To respect image licenses, we do not provide the original images shown to users, but instead provide embeddings for each image using DINOv2~\cite{oquab2023dinov2} with a ViT-B/14 backbone~\cite{dosovitskiy2020image}, as well as a ResNet50~\cite{He2015Dec} pretrained on ImageNet~\cite{Deng}.
For DINOv2, we average-pooled the final layer’s patch tokens after LayerNorm, excluding special tokens. 
For ResNet-50, we use the output of the global average pooling layer before the classifier.

To encode historical interactions, we construct a unique mapping for all species labels that appear in the dataset either as correct answers, or candidate options, and use it to encode the user's histories. In case of ``None of the above'' (NOTA) selections, the user's answer is encoded with a special token.

{\bf Dataset Characteristics.} \datasetname{} captures the learning dynamics of a diverse user population on a challenging real-world fine-grained classification task. 
It contains \totalRowsDataset{} user interactions, of which 98\% involve unique image-species-choice combinations, and 26\% contain unique species-choice pairs.
83\% of images are never seen more than once, resulting in 14,753,114 distinct image features using the ImageNet ResNet50. 
For DINOv2, we provide 14,747,840 features, the discrepancy (5,274) arising from the fact that DINOv2 was extracted later and images get deleted by the image owners over time. In our baselines, we treat unavailable images as zero inputs.

With over 10,000 species, \datasetname{} spans a broad taxonomic range and exhibits wide variation in user engagement. 
As shown in the first panel of \cref{fig:dataset_stats}, over 50\% of the 40,000+ users answered 100+ questions, and over 10\% exceed 1,000 questions, enabling the analysis of learning dynamics over time. 
Users encounter a mean of 138 distinct species, and 10\% encounter more than 300. 
Panel 2 and 3 of \cref{fig:dataset_stats} show that the distribution of accuracies across users and species is broad, and centered around 60-70\%. User improvement is observable, with over half of users showing measurable gains over sliding windows of 20 questions (one full quiz), making the dataset valuable for studying skill acquisition and feedback-driven learning. 
Panel 4 of \cref{fig:dataset_stats} also illustrates this trend, as within the first 20 times a user sees a particular species, their accuracy increases by 20\% on average. While focused on image-based identification, \datasetname{} also contains substantial geographic diversity, with quiz selections drawn from over 4,000 distinct locations (see \cref{fig:world_map} for a visualization), and temporal coverage spanning all weeks of the year. 

\begin{figure}[t]
\centering
\begin{subfigure}[t]{0.51\textwidth}
    \centering
    \includegraphics[width=\textwidth]{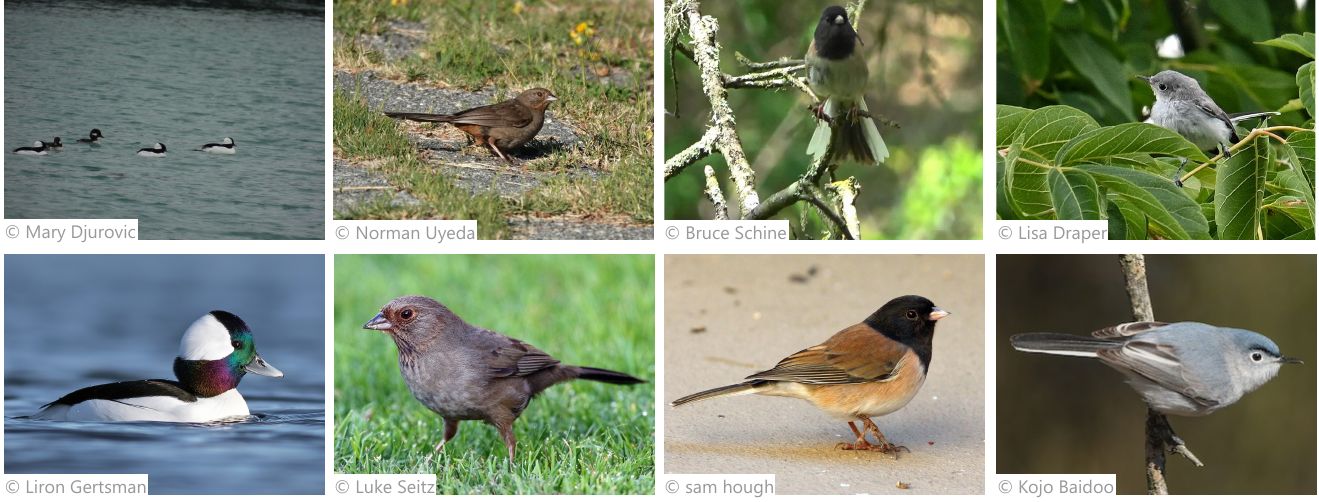}
\end{subfigure}
\hfill
\begin{subfigure}[t]{0.47\textwidth}
    \centering
    \includegraphics[width=\textwidth]{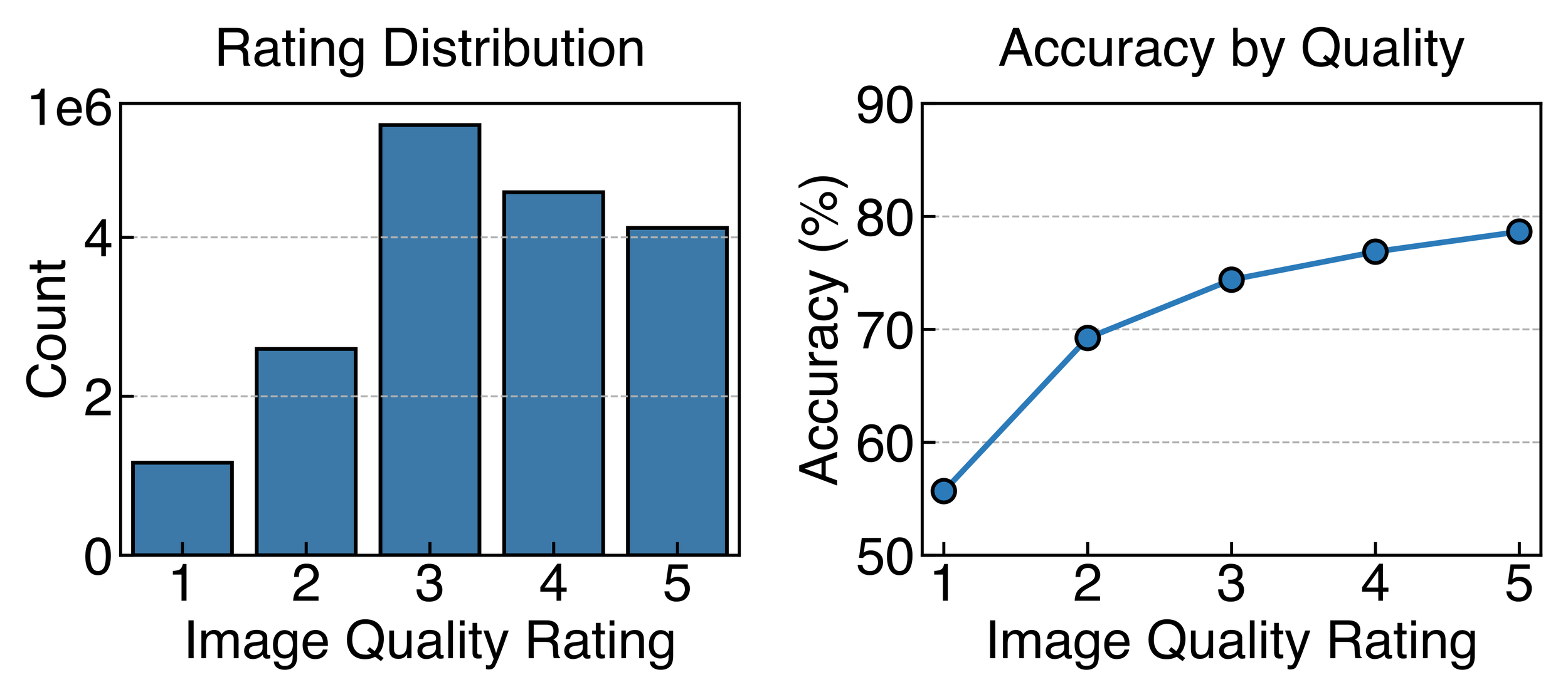}
\end{subfigure}
\caption{(Left) Here we compare  lower  quality quiz images (upper row) to high quality ones obtained from eBird species' pages (bottom row). Quiz questions may contain images that show birds from a distance, partially obscured, or uncommon angles. Species from left to right: Bufflehead, California Towhee, Dark-eyed Junco, and Blue-gray Gnatcatcher. 
(Right) Here we show the average accuracy of users for each possible quality rating. 
We observe that on average that higher quality images are easier for users. 
}
\label{fig:dataset_quality_stats}
\end{figure}

\begin{figure}[t]
\centering
\includegraphics[width=\textwidth]{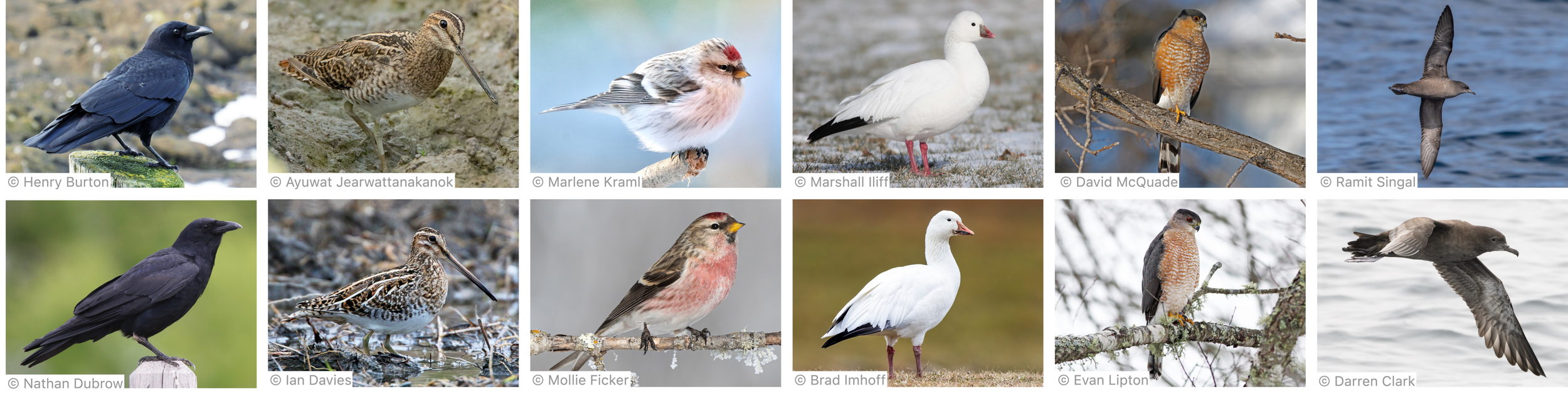}
\caption{Top-5 most frequently confused species pairs for species with > 1,000 interactions. From top-to-bottom and left-to-right: American Crow vs Fish Crow, Pin-tailed Snipe vs Common Snipe, Redpoll (Hoary) vs Redpoll (Common), Ross's Goose vs Snow Goose, Sharp-shinned Hawk vs Cooper's Hawk, and Short-tailed Shearwater vs Sooty Shearwater. Images taken from eBird~\cite{eBirdWeb}.}
\label{fig:confused_pairs}
\end{figure}

The difficulty of the task is further illustrated in \cref{fig:confused_pairs}, which displays the top-5 most frequently confused species pairs for species with over 1,000 interactions. For each species pair, the differences are subtle and require a trained eye to perceive. For example, the Pin-tailed Snipe can be differentiated from the Common Snipe by the white trailing edge of the wing of the Common Snipe~\cite{pintailedSnipe}. 
The Sharp-shinned Hawk can be differentiated from the Cooper's Hawk by its smaller head, more squared-off tail, and smaller feet~\cite{sharpshinnedhawk}. 
Note also, that the images shown for reference in \cref{fig:confused_pairs} are high quality example images of the species. In the quiz, users could be confronted with partially obscured or zoomed out images of the same bird, increasing difficulty. \cref{fig:dataset_quality_stats} shows example quiz images compared to high quality images of the same species, highlighting the inherent difficulty within the quiz. 
Nevertheless, users achieve an average accuracy of over 50\% even on images rated as low quality (see  \cref{fig:dataset_quality_stats} - right panel). 
The distribution of image quality is shown in \cref{fig:dataset_quality_stats} - center panel.

{\bf Privacy.} To ensure user privacy we anonymize all user-related identifiers, as well as those of quizzes, questions, and image assets that users interact with. We further aggregate quiz locations using the H3 geospatial index~\cite{hex3intro} at resolution 3~\cite{hex3GeoStats}, which averages \(12,393~\text{km}^2\) per cell. All quiz participants are registered Cornell Lab account holders and have agreed to the Terms of Use~\cite{ebirdTermsofUse}. The project has been reviewed and approved by the School of Informatics Ethics Committee (project number 954242).

\vspace{-5pt}
\section{Problem Setting}
\vspace{-5pt}

Our goal is to model the learning progress of human learners as they engage with a multi-choice image classification quiz. 
At time \( t \in \{1, \dots, T\} \), the learner is presented with an image \( I_t \) and an ordered list of \( K \) possible candidate answers, denoted by \(\mathbf{c}_t\). 
\(\mathbf{c}_t\) contains a set of $K-1$ randomly ordered possible candidate answers, and also includes an addition none of the above option \(\textsf{NOTA}\), yielding \(\mathbf{c}_t=(c_{t,1},\dots,c_{t,K-1},\textsf{NOTA})\).

We represent an image $I_t$ with a fixed vision encoder \( \mathbf{x}_t = f(I_t) \in \mathbb{R}^d\), and condition models on \(\mathbf{x}_t\).
The learner observes the image \( I_t \) and, based on their internal state, selects a response \( r_t \in \mathbf{c}_t\) to the question. 
The learner then receives the true species label \( y_t \) as feedback, and  proceeds to the next question in the quiz. 
This process forms a single interaction \( h_t = (\mathbf{x}_t, \mathbf{c}_t, y_t, r_t)\).

The learner's response is governed by their unobservable internal state \( \theta_{t} \), which summarizes their accumulated knowledge and memory, in conjunction with the input image and candidate choices shown. 
We assume the learner's state is updated after every interaction, similar to \cite{kondapaneni2022visual}, and model the learner’s response process as
\begin{equation}
r_t = \argmax_{c \in \mathbf{c}_t} P(c \mid \mathbf{x}_t, \mathbf{c}_t, \theta_t),
\end{equation}
where \( r_t \) denotes the categorical species selected by the learner and \( r_{t}^{\text{bin}} = \mathbb{I}[y_t = r_t] \in \{0,1\} \) indicates whether the answer was correct.
Here, \( P(\cdot) \) represents the learner’s true (but unknown) response distribution conditioned on their internal state. 

Given \( P(\cdot) \) is unobserved, we approximate it via a shared parametric model \( \phi \) trained across learners.
\( \phi \) does not include learner-specific parameters, instead, learner-specific behavior emerges through conditioning on each individual’s recent interaction history 
\( \mathcal{H}_t = (h_{\tau})_{\tau = \max(1,\, t - W)}^{t - 1}
\), along with the current question \((\mathbf{x}_t, \mathbf{c}_t, y_t)\). Here, \( W \in \mathbb{N} \) determines the maximum number of historical quiz questions that are assumed to influence the learner at a given time. While learners may in practice gain experience from other sources (\eg in the wild observations), we exclude such influences and assume a direct correspondence between a learner's state \( \theta_{t} \) and their interaction history \( \mathcal{H}_t\). For some models, the conditioning set is further augmented by including features such as the learner's chosen quiz location and time focus, or by removing information such as the image features.

The model \( \phi \), as an approximation of $P(\cdot)$,  estimates a response probability distribution and predicts either the categorical outcome as 
\begin{equation}
\hat{r}_t = \argmax_{c \in \mathbf{c}_t} \phi(c \mid \mathbf{x}_t, \mathbf{c}_t, y_t, \mathcal{H}_{t}),
\end{equation}
or binary outcome as \( \hat{r}_{t}^{\text{bin}} = \mathbb{I}\!\left[y_t = \hat{r}_t\right] \). 
The model is supervised to estimate either the response likelihood 
\( \phi(r_t \mid \mathbf{x}_t, \mathbf{c}_t, y_t, \mathcal{H}_{t}) \)
or, in the binary formulation, the correctness likelihood 
\( \phi(r_{t}^{\text{bin}} = 1 \mid \mathbf{x}_t, \mathbf{c}_t, y_t, \mathcal{H}_{t}) \).

{\bf Images.} As in \cite{kondapaneni2022visual}, we use a fixed vision encoder to represent images for modeling. 
Individual learners may weight or attend to different features within these embeddings.
We assume that learners first construct an internal visual representation of the image, which is subsequently used for species categorization, in line with studies indicating categorical readout from learned visual representations~\cite{ashby2011human, jiang2007categorization}.
Although this abstraction does not perfectly capture the learner's true visual representation, we hypothesize that human participants and neural networks trained on the same task extract similar image concepts.
Due to their high dimensionality and pretraining on real-world images, we expect these features to encode much of the information used by both novice and expert learners. 
Prior work has demonstrated that pretrained CNN features can be predictive of human similarity judgments~\cite{attarian2020transforming}.
We present results on the predictive accuracy of these image features for species classification in \cref{tab:image_classification}.

\section{Methods}
\label{sec:models}
\vspace{-5pt}

We quantitatively evaluate \datasetname{} across a range of different models \(\phi\) predicting user responses given different levels of context contained in the quiz question: correct species \(y_t\), choice candidates \(\mathbf{c}_t\), image features \(\mathbf{x}_{t}\), interaction history \(\mathcal{H}_t \), and focus indicators $i_{loc}$ and $i_{st}$. 
This context can be categorized into three main domains: 
(1) \emph{User Context}: Features that are directly associated to a particular user, such as transcripts from their interaction history, or their previous performance and preferences. This context is aggregated from the interaction history \(\mathcal{H}_t \) and focus indicators $i_{loc}$ and $i_{st}$. 
(2) \emph{Species Context}: Species-level features that are aggregated from the training set, such as average species difficulty of the correct species \(y_t\) and choice candidates \(\mathbf{c}_t\).
(3) \emph{Image Context}: Extracted image features, implicitly encoding image concepts such as quality and ambiguity. 
Models relying on user, species, and image context are denoted with \texttt{U}, \texttt{S}, and \texttt{Img} respectively. 
Combinations are denoted with \texttt{U+S} or \texttt{U+S+Img}, indicating user and species, or user, species, and image context, respectively.

{\bf Evaluation.} We evaluate the methods on a dataset of held out user IDs.
The evaluation metrics for the binary task are
binary macro accuracy which is the macro averaged accuracy over the correct outcome versus incorrect,
and binary AP for correctly predicting user mistakes which is the average precision with the minority incorrect class treated as the ``positive'' class.
For the multiple choice task, we report 
multiple choice accuracy which is accuracy over labels 1 to 5, where 5 denotes ``None of the above'',
and multiple choice incorrect set accuracy which is the accuracy computed on the subset of questions that participants answered incorrectly, measuring how well the model recovers the correct label among distractors for questions participants fail to answer correctly. 

{\bf Multiple-choice Classifiers.} For multiple-choice response prediction, we evaluate transformer-based KT models, a confusion prior classifier, a simple MLP, and two heuristics. 
The first heuristic assumes an all-knowing learner that always selects the correct answer (\textit{Always Correct}). 
The confusion prior classifier (\textit{Conf Prior}) estimates the probability of each choice by masking the training-set confusion between the correct species and distractors, then re-normalizing to form a valid distribution over the choices presented to the user.
We also add an additional confusion prior model which is constrained to predict only incorrect choices. This setup simulates a confusion prior focused exclusively on the subset of user questions answered incorrectly.
As a lightweight neural baseline, we train a one-layer MLP that receives a learned 250-dimensional embedding of the correct species along with optional context (none, user-context, species-context, or both). 
The model outputs a probability distribution over all species, which is then masked to include only the five presented choices. We test user and species context (\textit{MLP} \texttt{U+S}) and user, species, and image context (\textit{MLP} \texttt{U+S+Img}). 
The image features are passed through an embedding layer converting them to the MLP input dimension.  
For both the confusion prior classifier and the one-layer MLP, NOTA is selected if the total probability assigned to species outside the available options exceeds the probability of every presented choice.

{\bf Binary Classifiers.} For correct/incorrect prediction we fit simple probabilistic models (logistic regression (\textit{LR}), XGBoost (\textit{XG}), random forests (\textit{RF}) with combinations of user and species context (\texttt{U}, \texttt{S}, and \texttt{U+S}). Additionally, we fit an average species classifier heuristic (\textit{Avg Species}), which mirrors average species accuracy in the training set. 
A full description of models and their hyperparameters can be found in \cref{appendix:implementation_details}.

{\bf Knowledge Tracing Models.} We also evaluate several \emph{knowledge tracing} baselines on our binary classification task. simpleKT~\cite{Liu2023Feb} and Knowledge Query Networks (KQN)~\cite{Lee2019Mar} serve as lightweight baselines. DKT~\cite{piech2015deep}, its regularized variant DKT+~\cite{yeung2018addressing}, and the adversarial-training-based knowledge tracing (ATKT)~\cite{Guo2021Oct}, which adds adversarial training, model a learner’s history using an LSTM hidden state. The self-attentive model for knowledge tracing (SAKT)~\cite{pandey2019self} and attentive knowledge tracing (AKT)~\cite{ghosh2020context} selectively attend to the most relevant past interactions, with AKT using Rasch model~\cite{Rasch1993} inspired regularization. Dynamic Key-Value Memory Networks for Knowledge Tracing (DKVMN)~\cite{zhang2017dynamic} externalizes knowledge into a key–value memory, enabling long-range tracking of per-concept knowledge.

We also evaluate two language modeling training paradigms for our multiple-choice task: sequence-to-sequence (LM-Seq2seq) generation~\cite{Raffel2020} and multiple-choice classification (LM-MCC)~\cite{zellers-etal-2018-swag}. 
For seq2seq, we fine-tune a pretrained T5-style \cite{Raffel2020} encoder-decoder transformer \cite{Tay2021Sep} to generate the correct species token based on the user's interaction history and the current question. 
For the classification model, we fine-tune a pretrained Bert-style \cite{Devlin2019Jun} encoder-only transformer \cite{Jiao2019Sep} to score the compatibility of each question-choice pair, conditioned on the same history.

For both models, we use a custom tokenizer which only includes one token per possible species, \ie \uniquespeciesnumIcnlChoices{} tokens, along with special tokens for padding, segment separation, and token types.
This avoids ambiguity when using natural language to describe species names, and ensures a compact token representation of our task, with each question occupying exactly 8 tokens. 

For the seq2seq model, supervision is applied via a token-level cross-entropy loss on the generated species token, with NOTA aggregating all probabilities outside the visible choice candidates. 
For the classification model, binary labels are assigned to each candidate, with exactly one correct answer per question. 
For these transformer models, we set \(W=50\),  
other models use the complete history. 
These transformer models are evaluated using accuracy on the full dataset and subset of incorrect answers, as in case of the other multiple-choice models.
Please see \cref{appendix:implementation_details} for additional implementation details for these models.

{\bf Image Features.} To showcase the use of image features for our prediction task, we evaluate an MLP with image features as input, with and without additional user and species context. The MLP using only image features (\textit{MLP} \texttt{Img}), omitting user and species context,  is similar to the static tracing model in~\cite{kondapaneni2022visual}. Image features are encoded via an embedding layer to a dimension matching the number of unique species in the dataset. We report results for DINOv2 image features below, but full results, including using ResNet-50 image features, can be found in \cref{tab:grouped_combined_metrics}.

\section{Results}
\label{sec:results}
\vspace{-5pt}

\begin{figure}[t]
\centering
\includegraphics[width=\textwidth]{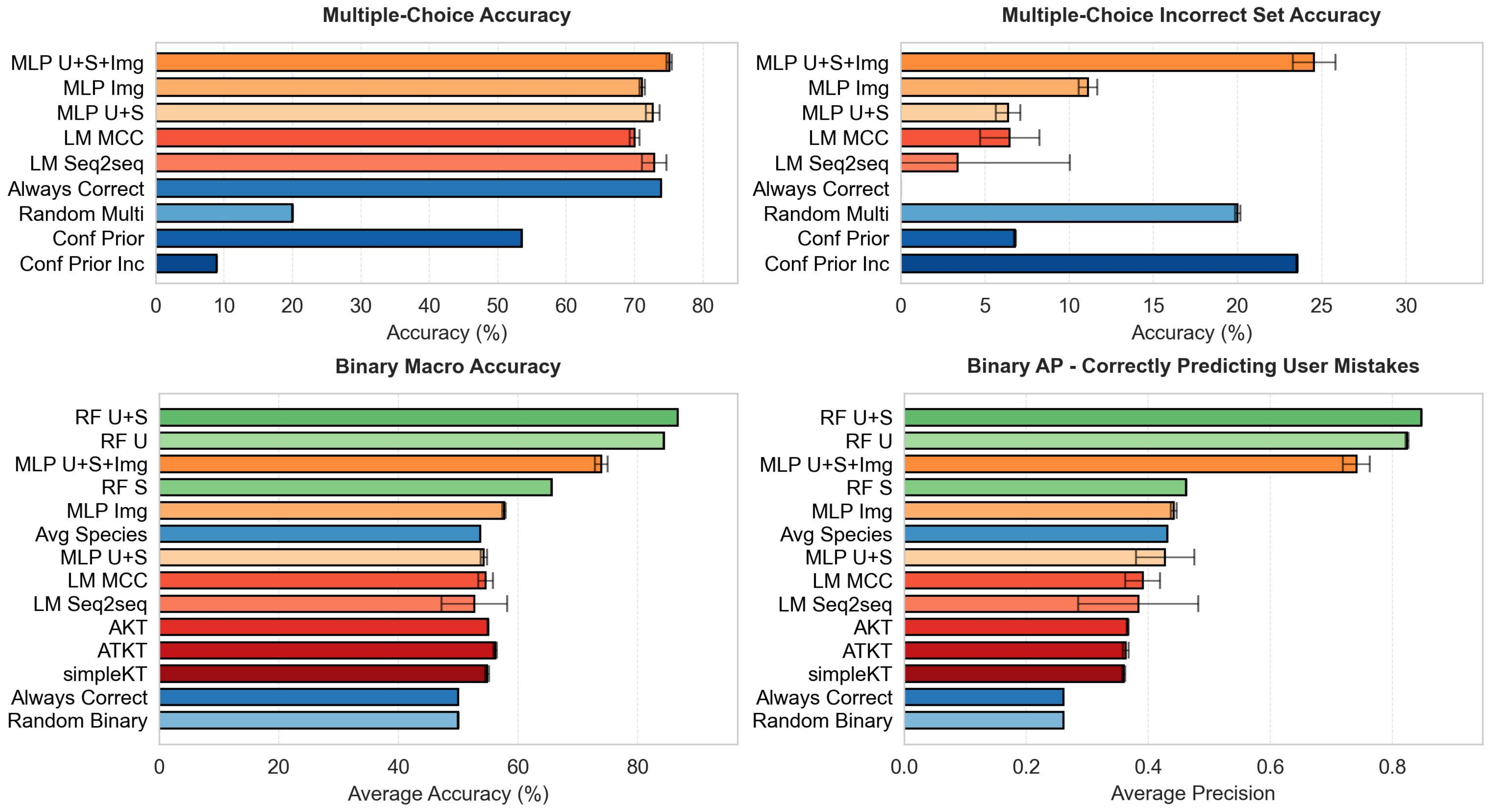}
\setlength{\abovecaptionskip}{-3pt}
\setlength{\belowcaptionskip}{-15pt}
\caption{Performance on the multiple-choice and binary tasks. Top-left: accuracy on the full multiple-choice dataset. Top-right: accuracy on the subset of questions answered incorrectly. 
Bottom-left: macro-averaged accuracy on the binary task. 
Bottom-right: average precision (AP) for predicting user errors. 
Models are grouped by color into \textcolor{mydarkgreen}{simple classifiers} (\textit{RF} \texttt{U}, \textit{RF} \texttt{S}, \textit{RF} \texttt{U+S}), \textcolor{orange}{MLPs} (\textit{MLP} \texttt{U+S+Img}, \textit{MLP} \texttt{Img}), \textcolor{myred}{KT models} (\textit{LM MCC}, \textit{LM Seq2seq}, \textit{AKT}\cite{ghosh2020context}, \textit{ATKT}\cite{Guo2021Oct}, and \textit{simpleKT}~\cite{Liu2023Feb}), and simple \textcolor{blue}{heuristics} (\textit{Always Correct}, \textit{Random binary}, \textit{Random multiple-choice}, \textit{Conf Prior}, \textit{Conf Prior Inc}). \textit{Img} uses DINOv2 features here, full results can be found in \cref{app:add results}.}
\label{fig:results}
\end{figure}

We focus our discussion on high level takeaways from~\cref{fig:results}. 
Additional results (\eg ResNet-50 image feature results, and additional KT baseline results) can be found in the appendix.

{\bf Feature engineered context result in strong baselines.}
The bottom row of~\cref{fig:results} shows performance on the binary classification task, where the goal is to predict whether a participant will correctly answer a given question.  
Models trained specifically for this binary task outperform those trained for multiple choice and then subsequently binarized.
The random forest model, leveraging both user and species context, performs best overall, achieving over 80\% average precision in predicting participant errors.  
On the multiple-choice task, a one-layer MLP with user and species context matches the performance of significantly larger transformer models with a much larger capacity.

{\bf User context is more informative than species context.} 
When comparing different context types, models receiving both user and species context  (\ie \texttt{U+S}) perform best overall, with models receiving only user context (\ie \texttt{U})  close behind.  
Among the binary classifiers, those incorporating user context consistently achieve higher accuracy and average precision.  
While species context models still outperform simple heuristics on the binary task and full-dataset multiple-choice task, user-specific context appears necessary for strong predictions.

{\bf Multiple-choice trained classifiers are beaten by binary-trained classifiers on the binary task.}
On the binary task, multiple-choice classifiers are consistently outperformed by those trained directly on the binary objective. Despite the small capacity of our probabilistic models, they can achieve an AP and average accuracy of over 80\%.
By contrast, models originally trained to predict the exact multiple-choice response must allocate capacity not only to correctness but also to the structure of the incorrect responses.  
Future work could explore how these multiple-choice models could benefit from the outputs of binary classifiers, either as auxiliary signals or as gating mechanisms, or receive additional binary supervision.

{\bf Image features complement user and species context for incorrect choices.} On the full dataset, the MLP with user and species context and DINOv2 image feature input shows a slightly higher performance compared to the other multiple-choice trained models (76\%). On only incorrectly answered questions, however, it significantly outperforms the other parametric models and the random baseline with results around 25\%. The same MLP without user and species context, \textit{MLP} \texttt{Img}, achieves comparable results on the full dataset, but achieves only around 11\% on the incorrect subset. This shows that image features help knowledge tracing for \datasetname{}, especially if combined with the right context.

{\bf Predicting incorrect choices is challenging.} Our trained models achieve approximately 70\% accuracy on the multiple-choice task (see~\cref{fig:results}, top left panel).  
However, as observed in \cref{fig:dataset_stats}, participants perform substantially above random chance.
The top right panel of~\cref{fig:results} shows model accuracy on the subset of questions where participants select the \emph{incorrect} answer.
All trained models achieve less than 25\% accuracy on this subset, while the random classifier attains the expected 20\% (\ie a one-in-five chance), and the always-correct classifier scores 0\% by design.
The confusion prior, which assumes users are always incorrect (\textit{Conf Prior Inc}), predictably underperforms on the full dataset (below 9\% accuracy), but achieves over 23\% accuracy on the incorrect subset.
High performance on this subset would indicate that a model is able to approximate the participants’ internal knowledge states.
The marginal improvement of the best trained model over the incorrect-assumption confusion prior suggests that there is substantial room for improvement.
Future work could explore the use of  longer temporal contexts and better generalization across species, since success or failure on one species could provide information for others.

{\bf Species context can improve incorrect predictions.} To isolate whether species context can guide incorrect predictions even without image features, we created an additional confusion prior classifier, \textit{Conf Prior Inc}, and restricted it to never answer the correct species. As expected, this classifier performs poorly on the full dataset (accuracy of ~10\%). On the incorrect subset, it managed to outperform the random baseline, and obtain a similar accuracy as the MLP classifier with user, species, image context (around 24\%). This shows that species context can help with predicting incorrect guesses.

{\bf Knowledge tracing baselines have little variance in performance.}
~\cref{tab:pykt_results_full} summarizes the performance of the KT methods. As in~\cite{liu2023xes3g5m}, the tested KT methods all show similar performance, with average precision on predicting user's mistakes around 0.35 and average accuracy around 54\%. This is surprising, given the diversity of model architectures. In terms of average accuracy, these models underperform compared to our simple classifiers incorporating user and species context.
We note that most KT methods are designed for datasets with far fewer concepts than \datasetname{}, and typically offer concept-level interpretability, which can limit their flexibility. As shown in \cref{fig:results}, macro accuracy and AP (mistakes) are low for the KT methods AKT, ATKT, and simpleKT, despite high accuracy on the positive class. Similar results are shown for additional KT methods in \cref{tab:pykt_results_full}. This suggests suboptimal performance on the negative class, which are instances where participants selected an incorrect species. One plausible explanation is that these models exploit a shortcut, focusing disproportionately on the positive class. We view this as evidence of CleverBirds’ value in exposing limitations of existing approaches which motivates the development of more effective methods in future.

\section{Limitations}
\label{sec:limitations}
\vspace{-5pt}

While \datasetname\ is the largest and most diverse dataset for visual knowledge tracing, it has some limitations.  
First, it focuses exclusively on bird species identification.  
We acknowledge this domain specificity limits immediate applicability to other areas such as medical imaging or object recognition. 
This is partially mitigated by the visual diversity of birds and the prevalence of hard-to-distinguish species pairs, making it representative of many fine-grained classification tasks. 
Moreover, the dataset's scale of nearly 11,000 fine-grained categories and over 40,000 participants with extensive learning trajectories provides unprecedented insights into human visual learning dynamics that, while domain-specific, may inform modeling approaches across fine-grained recognition tasks. 

Second, the dataset reflects both location and selection biases.  
Participants are drawn from eBird~\cite{sullivan2009ebird}, and participants on such platforms are known to be skewed towards the Global North~\cite{daru2023mass}. See \cref{app:geogr_bias} for additional details on geographic bias. 
The multiple choice format is a tradeoff that differs from open-ended identification, limiting knowledge tracing to a finite set of options and influencing ecological validity. This is partially mitigated through controlled design of the quiz, \ie distractors are dynamically sampled from a sliding window centered on the true species, ensuring variability and substantial difficulty. Distractors are selected to be taxonomically similar, making the task challenging and realistic. 
A related constraint is the feedback, which only provides the correct species label to the user. Despite this simplicity, the signal is highly effective for learning in our fine-grained setting. Participant accuracy improves significantly with repeated species exposure (\cref{fig:dataset_stats}), showing that the correct label is a potent signal for correcting subtle classification errors. This finding is consistent with prior work showing humans can acquire visual expertise from label supervision alone \citep{singla2014near}.

Finally, although some label noise may be present in the quiz answers, we do not expect it to be substantial (\cref{sec:dataset}), and the primary task is to predict user responses rather than the ground truth species labels.
As with all KT applications, care must be taken when developing models from datasets such as ours, as inaccurate models could negatively bias future human learning.

\section{Conclusion}
\vspace{-5pt}

We introduce \datasetname, a new benchmark for evaluating models on the task of fine-grained visual knowledge tracing. 
\datasetname contains rich interaction data from over 40,000 participants who are attempting to visually discriminate over 10,000 different bird species from all over the world. 
There are several properties that make our dataset well suited as a benchmark for this task, \eg it contains a large number of interactions over time originating from participants of different skill levels, participants are not static as we observe their overall performance improves over time, and the concept space is large and challenging to master. 
We also demonstrate that \datasetname poses a challenge for the computational methods tested, but the inclusion of additional context information improves performance. 
\datasetname opens the door to future avenues related to modeling human knowledge acquisition in complex real world visual discrimination tasks, in addition to spurring development of methods for teaching such knowledge to learners.

\vspace{10pt}
\noindent{\bf Acknowledgments.} 
We thank the eBird users for providing bird images, the Macaulay Library for curating and hosting this content, and the eBird participants who took part in the Photo and Sound Quiz.
Media from the Cornell Lab of Ornithology | Macaulay Library was embedded as image features, published, used for baselines, and a subset was included for illustration in the paper.
LB was supported by the United Kingdom Research and Innovation (grant EP/S02431X/1), UKRI Centre for Doctoral Training in Biomedical AI at the University of Edinburgh, School of Informatics. 
OMA was in part supported by a Royal Society Research Grant.

\clearpage
\newpage

\bibliographystyle{abbrvnat} %
\bibliography{main}

\newpage
\include{checklist}

\appendix
\setcounter{table}{0}
\renewcommand{\thetable}{A\arabic{table}}
\setcounter{figure}{0}
\renewcommand{\thefigure}{A\arabic{figure}}
\noindent{\bf\LARGE Appendix}\\

\section{Additional Results}
\label{app:add results}

\subsection{Full Results}

\cref{tab:grouped_combined_metrics} and \cref{tab:pykt_results_full} report results on \datasetname{}. Complementing \cref{fig:results}, \cref{tab:grouped_combined_metrics} includes models using ResNet-50 image features (\textit{MLP} \texttt{Img ResNet-50} and \textit{MLP} \texttt{U+S+Img Resnet-50}), logistic regression (\textit{LR} \texttt{U}, \textit{LR} \texttt{S}, \textit{LR} \texttt{U+S}) and XGBoost (\textit{XG} \texttt{U+S}). 
\cref{tab:pykt_results_full} reports additional results for the KT baselines AKT~\cite{ghosh2020context}, simpleKT~\cite{Liu2023Feb}, DKT~\cite{piech2015deep}, DKT+~\cite{yeung2018addressing}, DKVMN~\cite{zhang2017dynamic}, KQN~\cite{Lee2019Mar}, ATKT~\cite{Guo2021Oct}, SAKT~\cite{pandey2019self}, SKVMN~\cite{abdelrahman2019knowledge} on the binary task, evaluated using \textit{AP (mistakes)}, \textit{Macro Accuracy (\%)}, \textit{AUC} and \textit{Accuracy (\%)}. 
\textit{AP (mistakes)} and \textit{Macro Accuracy (\%)} are directly comparable with columns 2 and 3 of \cref{tab:grouped_combined_metrics}, while \textit{AUC} and \textit{Accuracy (\%)} are directly comparable with Table 2 of \cite{liu2023xes3g5m}. Both our results and those of \cite{liu2023xes3g5m} show little variability across model types.

\begin{table}[h]
\centering
\caption{Baseline results on the multiple-choice and binary tasks. Columns (2) and (3) show performance on the binary classification task (correct/incorrect). Columns (4) and (5) show performance on the multiple-choice classification task. AP is average precision on predicting learners' mistakes, macro accuracy is average accuracy over the two classes (macro recall). `Full' describes the full dataset, `Incorrect' only incorrectly-answered questions. 
Models are: (1) Knowledge tracing models: sequence-to-sequence transformer model (\textit{LM Seq2seq}), multiple-choice classification transformer model (\textit{LM MCC}), \textit{AKT}~\cite{ghosh2020context}, \textit{ATKT}~\cite{Guo2021Oct} and \textit{simpleKT}~\cite{Liu2023Feb}; (2) MLPs using user context (U),  species context (S), and image features from ResNet-50 (\textit{Img ResNet-50}) and DINOv2 (\textit{Img DINOv2}) features; (3) Random Forest (\textit{RF}), Logistic Regression (\textit{LR}) and XGBoost (\textit{XG}); (4) Random binary and multiple-choice baselines, and heuristics using the average species performance (\textit{Avg Species}), user-correct assumption (\textit{Always Correct}), confusion prior (\textit{Conf Prior}) and the confusion prior which assumes incorrect participant guesses (\textit{Conf Prior Inc}). 
Error bars are 2-sigma, based on 3 training runs for KT baselines, and 10 training runs for all other models. Metrics are reported as averages, with 2-sigma standard deviation.
}
\vspace{5pt}
\resizebox{1.0\textwidth}{!}{
\begin{tabular}{l c c c c}
\toprule
Task & \multicolumn{1}{c}{\textbf{Binary}} & \multicolumn{1}{c}{\textbf{Binary}} & \multicolumn{1}{c}{\textbf{Multiple Choice}} & \multicolumn{1}{c}{\textbf{Multiple Choice}} \\
Metric & \multicolumn{1}{c}{\textbf{AP}} & \multicolumn{1}{c}{\textbf{Macro Accuracy (\%)}} & \multicolumn{1}{c}{\textbf{Accuracy (\%)}} & \multicolumn{1}{c}{\textbf{Accuracy (\%)}} \\
Dataset & \multicolumn{1}{c}{\textbf{Full}} & \multicolumn{1}{c}{\textbf{Full}} & \multicolumn{1}{c}{\textbf{Full}} & \multicolumn{1}{c}{\textbf{Incorrect}} \\
\midrule
LM Seq2seq & 0.384 $\pm$ 0.098 & 66.90 $\pm$ 21.60 & 72.90 $\pm$ 1.80 & 3.40 $\pm$ 6.60 \\
LM MCC & 0.391 $\pm$ 0.028 & 66.60 $\pm$ 18.20 & 70.00 $\pm$ 0.60 & 6.50 $\pm$ 1.80 \\
AKT & 0.366 $\pm$ 0.002 & 55.00 $\pm$ 0.00 & -- & -- \\
ATKT & 0.363 $\pm$ 0.004 & 56.20 $\pm$ 0.40 & -- & -- \\
simpleKT & 0.360 $\pm$ 0.002 & 54.80 $\pm$ 0.40 & -- & -- \\
\midrule
MLP U+S & 0.428 $\pm$ 0.048 & 67.60 $\pm$ 19.20 & 72.70 $\pm$ 1.00 & 6.40 $\pm$ 0.80 \\
MLP U+S+Img ResNet-50 & 0.768 $\pm$ 0.042 & 79.90 $\pm$ 9.60 & \textbf{76.00 $\pm$ 1.20} & \textbf{24.50 $\pm$ 2.40} \\
MLP U+S+Img DINOv2 & 0.741 $\pm$ 0.022 & 79.90 $\pm$ 8.80 & 75.10 $\pm$ 0.40 & \textbf{24.50 $\pm$ 1.20} \\
MLP Img ResNet-50 & 0.461 $\pm$ 0.004 & 68.60 $\pm$ 18.00 & 72.90 $\pm$ 0.20 & 9.00 $\pm$ 0.40 \\
MLP Img DINOv2 & 0.442 $\pm$ 0.004 & 68.70 $\pm$ 15.80 & 71.10 $\pm$ 0.40 & 11.10 $\pm$ 0.60 \\
\midrule
RF U & 0.824 $\pm$ 0.002 & 81.40 $\pm$ 4.20 & -- & -- \\
RF S & 0.462 $\pm$ 0.000 & 64.80 $\pm$ 1.20 & -- & -- \\
RF U+S & \textbf{0.848 $\pm$ 0.000} & 84.90 $\pm$ 2.60 & -- & -- \\
LR U & 0.809 $\pm$ 0.002 & 80.60 $\pm$ 5.20 & -- & -- \\
LR S & 0.433 $\pm$ 0.000 & 64.70 $\pm$ 1.00 & -- & -- \\
LR U+S & 0.824 $\pm$ 0.002 & 83.80 $\pm$ 3.40 & -- & -- \\
XG U+S & 0.845 $\pm$ 0.000 & \textbf{85.40 $\pm$ 4.00} & -- & -- \\
\midrule
Random Binary & 0.261 $\pm$ 0.000 & 50.00 $\pm$ 0.00 & -- & -- \\
Random Multi & 0.261 $\pm$ 0.000 & 40.40 $\pm$ 13.60 & 20.00 $\pm$ 0.00 & 20.00 $\pm$ 0.20 \\
Avg Species & 0.432 $\pm$ 0.000 & 67.60 $\pm$ 19.80 & -- & -- \\
Always Correct & 0.261 $\pm$ 0.000 & 65.90 $\pm$ 22.80 & 73.90 $\pm$ 0.00 & 0.00 $\pm$ 0.00 \\
Conf Prior & 0.252 $\pm$ 0.000 & 55.50 $\pm$ 9.80 & 53.50 $\pm$ 0.00 & 6.80 $\pm$ 0.00 \\
Conf Prior Inc & 0.245 $\pm$ 0.000 & 34.70 $\pm$ 20.80 & 8.90 $\pm$ 0.00 & 23.50 $\pm$ 0.00 \\
\bottomrule
\end{tabular}
}
\vspace{0.5em}
\label{tab:grouped_combined_metrics}
\end{table}

\begin{table}[ht]
\centering
\caption{Results of knowledge tracing models on the binary classification task. Metrics are \textit{AP}, \textit{macro accuracy (\%)}, \textit{AUC} and \textit{accuracy (\%)}. Averages are reported with 2-sigma standard deviation on 3 runs. The models receive user context and predict binary correct/incorrect outcomes, the AP and Macro Accuracy values can thus be compared to columns (2) and (3) of~\cref{tab:grouped_combined_metrics}.}
\vspace{5pt}
\resizebox{0.85\columnwidth}{!}{%
\begin{tabular}{lllll}
\toprule
\textbf{Model} & \textbf{AP (mistakes)} & \textbf{Macro Accuracy (\%)} & \textbf{AUC} & \textbf{Accuracy (\%)} \\
\midrule
AKT~\cite{ghosh2020context} & \textbf{0.366} $\pm$ 0.001 & 54.976 $\pm$ 0.068 & 0.632 $\pm$ 0.002 & 72.293 $\pm$ 0.115 \\
simpleKT~\cite{Liu2023Feb} & 0.360 $\pm$ 0.002 & 54.803 $\pm$ 0.262 & 0.628 $\pm$ 0.001 & 72.034 $\pm$ 0.013 \\
DKT~\cite{piech2015deep} & 0.334 $\pm$ 0.002 & 54.137 $\pm$ 0.113 & 0.600 $\pm$ 0.001 & 70.369 $\pm$ 0.118 \\
DKT+~\cite{yeung2018addressing} & 0.333 $\pm$ 0.000 & 53.771 $\pm$ 0.000 & 0.600 $\pm$ 0.000 & 70.620 $\pm$ 0.000 \\
DKVMN~\cite{zhang2017dynamic} & 0.349 $\pm$ 0.003 & 54.941 $\pm$ 0.137 & 0.613 $\pm$ 0.002 & 71.017 $\pm$ 0.337 \\
KQN~\cite{Lee2019Mar} & 0.340 $\pm$ 0.000 & 54.438 $\pm$ 0.297 & 0.604 $\pm$ 0.004 & 70.694 $\pm$ 0.505 \\
ATKT~\cite{Guo2021Oct} & 0.363 $\pm$ 0.004 & \textbf{56.183} $\pm$ 0.248 & \textbf{0.636} $\pm$ 0.004 & 70.977 $\pm$ 0.229 \\
SAKT~\cite{pandey2019self} & 0.342 $\pm$ 0.002 & 54.468 $\pm$ 0.185 & 0.604 $\pm$ 0.002 & 71.103 $\pm$ 0.224 \\
SKVMN~\cite{abdelrahman2019knowledge} & 0.350 $\pm$ 0.001 & 52.920 $\pm$ 0.229 & 0.608 $\pm$ 0.001 & \textbf{73.138} $\pm$ 0.174 \\
\bottomrule
\end{tabular}
}
\label{tab:pykt_results_full}
\end{table}

\begin{table}[t]
   
    \centering %
    \caption{Comparison of existing knowledge tracing datasets.  Table extended from~\cite{kim2024kt}.
    }
   \resizebox{1\columnwidth}{!}{
   \begin{tabular}{@{}l rrrr l @{}}
        \toprule
        Dataset & Participants &  Questions & Concepts &  Interactions & Subject \\
        \midrule
        Simulated-5~\cite{piech2015deep} & 4,000 &  50& 5& 200,000 & Synthetic\\
        ASSISTments2009~\cite{feng2009addressing} & 4,217 & 26,688 & 123 & 346,860 & Math \\
        ASSISTments2012~\cite{feng2009addressing} & 46,674 & 179,999 & 265 & 6,123,270 & Math \\
        ASSISTments2015~\cite{feng2009addressing} & 19,917 & 100 & - & 708,631 & Math \\
        ASSISTments2017~\cite{feng2009addressing} & 1,709 & 3,162 & 102 & 942,816 & Math \\
        Statistics2011~\cite{olilearning} & 333 & 1,224 & - & 194,947 & Math \\
        Junyi2015~\cite{chang2015modeling} & 247,606 & 722 & 41 & 25,925,922 & Math \\
        KDD2005~\cite{stamper2010kddcup2005} & 574 & 210,710 & 112 & 809,694 & Math \\
        KDD2006~\cite{stamper2010kddcup2006} & 1,146 & 207,856 & 493 & 3,679,199 & Math \\
        XES3G5M~\cite{liu2023xes3g5m} & 18,066 & 7,652 & 865 & 5,549,635 & Math \\
        Eedi Task 1~\cite{wang2020diagnostic} & 118,971 & 27,613 & - & 15,867,850 & Math \\
        Eedi Task 2~\cite{wang2020diagnostic} & 4,918 & 948 & - & 1,382,727 & Math \\
        ES-KT-24~\cite{kim2024kt} & 15,032 & 182  & 28 & 7,783,466 & Math and language \\
        POJ~\cite{pandey2020rkt} & 22,916 & 2,750 & - & 996,240 & Programming \\
        PTADisc~\cite{hu2023ptadisc} & 1,530,100 & 225,615 & 4,054 & 680M+ & Programming \\
        Programming~\cite{fu2024sinkt} & 2,756 & 726 & 82 & 193,284 & Programming \\
        EdNet~\cite{choi2020ednet} & 1,677,583 & 52,676 & 962 & 372,366,720 & Linguistics \\
        DBF-KT22~\cite{abdelrahman2022dbe} & 1,361 & 212 & 98 & 167,222 & Computer and information science \\ \hline 
        GravitySpy~\cite{crowston2019knowledge} & 10,655 & 51,047 & 21 & 1,026,652 & Spectrogram classification \\ 
        VTK-Greebles~\cite{kondapaneni2022visual} & 150 & 1,200 & 3 & 6,750 & Synthetic image classification \\ 
        VTK-Eyes~\cite{kondapaneni2022visual}  & 150 & 600 & 3 & 6,750 & Retinal disease classification  \\ 
        VTK-Butterflies~\cite{kondapaneni2022visual} & 150 & 2,224 & 5 & 6,750 & Butterfly species classification  \\
        {\bf \datasetname (Ours)} &\numusers  & \totalRowsDataset & \uniquespeciesnum & \totalRowsDataset & Birds species classification \\
        \bottomrule
    \end{tabular}
    }
    \label{tab:kt_datasets}
\end{table}

\subsection{In-Context Learning}\label{app:in_context}

To evaluate whether large language models can solve the task in-context, we evaluated OpenAI’s GPT-4.1 Nano~\cite{BibEntry2025Sep} and DeepSeek’s DeepSeek-V3-0324~\cite{deepseekai2024deepseekv3technicalreport} on the multiple-choice task, using a subset of 100,000 test set examples.
Each model received a natural language task description, the participant’s interaction history, and a prompt for classification. \cref{tab:prompt_structure} shows the exact prompts used. We use [HISTORY] to denote the participant’s interaction history, formatted for sequence-to-sequence prediction and rendered with actual species names in text rather than tokenized codes. The history sequence ends with the current question, encouraging the model to input the participant's answer. 

The results (\cref{tab:in_context_performance})  broadly reflect those of the fine-tuned models, i.e., GPT-4.1 Nano achieves high accuracy on correctly answered examples but struggles on incorrect ones, whereas DeepSeek exhibits lower overall accuracy but outperforms on the subset of incorrect examples.
Comparing this to the other models in \cref{fig:results}, GPT-4.1 Nano matches LM MCC on the full set, but underperforms the parametric models on the subset of incorrectly-answered questions. Deepseek-V3-0324 underperforms the other language models on the full test set, but outperforms them on the incorrectly-answered questions. We note two key limitations: we only evaluate a random subset of 100,000 examples, and we use identical prompts for both models without model-specific optimization.

\begin{table}[ht]
\centering
\caption{In-context multiple-choice accuracy on full dataset and incorrect subset using pre-trained cloud models for a random subset of 100,000 examples.}
\vspace{5pt}
\resizebox{0.5\columnwidth}{!}{%
\begin{tabular}{l c c}
\toprule
\textbf{Task} & \multicolumn{2}{c}{\textbf{Multiple Choice}} \\
\textbf{Metric} & \multicolumn{2}{c}{\textbf{Accuracy}} \\
\textbf{Dataset} & \multicolumn{1}{c}{\textbf{Full}} & \multicolumn{1}{c}{\textbf{Incorrect}} \\
\midrule
Count & \multicolumn{1}{c}{100,000} & \multicolumn{1}{c}{26,115} \\
\midrule
deepseek-chat~\cite{deepseekai2024deepseekv3technicalreport} & 0.53 & \textbf{0.13} \\
gpt-4.1-mini-2025-04-14~\cite{BibEntry2025Sep} & \textbf{0.71} & 0.02 \\
\bottomrule
\end{tabular}
}
\label{tab:in_context_performance}
\end{table}

\begin{table}[ht]
\centering
\caption{Prompts used for in-context learning task. [HISTORY] is replaced with the concatenation of all historical questions and the current question.}
\vspace{5pt}
\resizebox{1\columnwidth}{!}{%
\begin{tabular}{p{3cm}p{11cm}}
\toprule
\textbf{Component} & \textbf{Content} \\
\midrule
\textbf{System Prompt} & You are analyzing a user's quiz-taking behavior. In each quiz question, the user is shown an image and must choose the correct species from multiple-choice options. There are always five choices available: Four eBird species options, listed in the 'Options' field (indices 0–3). A fifth option labeled 'None of the Above', which is always choice 4. If the correct species is not among the four listed options, then 'None of the Above' (index 4) is the correct choice. Note: The 'None of the Above' option is always available but not shown in the 'Options' list. [HISTORY]. Your prediction should consider both the user’s interaction history and the difficulty of the current and previous questions. Use your knowledge of bird species when relevant. Only reply with the correct digit of the choice (0, 1, 2, 3, or 4). Your answer (single digit only): \\
\midrule
\textbf{Historical Questions} & Question: \\
& Correct answer: Northern Mockingbird \\
& Options: Northern Mockingbird | Eastern Bluebird | Western Bluebird | Mountain Bluebird \\
& User's answer: Northern Mockingbird \\
\midrule
\textbf{Current Question} & Question: \\
& Correct answer: Woodhouse's Scrub-Jay \\
& Options: Canada Jay | Pinyon Jay | Steller's Jay | Blue Jay \\
& User's answer: \\
\bottomrule
\end{tabular}
}
\label{tab:prompt_structure}
\end{table}

\subsection{Performance Inside and Outside the US}\label{app:geogr_bias}

To further analyse potential geographic biases, we compare performance across quizzes that were selected by participants as either inside or outside the US. 
We compare both the performance of participants on the quiz, as well as the knowledge tracing capability of our trained models. 

In \cref{tab:location_stats} we observe that participants exhibit no substantial performance difference when answering quizzes based on locations inside the US or outside it. 
In contrast, \cref{tab:combined_location_performance} shows that our models show a small but statistically significant performance gap favoring quizzes where the quiz location is in the US.

\begin{table}[ht]
\centering
\caption{Comparison of participant's performance on quiz locations inside and outside the US. Error bars are 2-sigma.}
\vspace{5pt}
\resizebox{0.4\columnwidth}{!}{%
\begin{tabular}{l c c}
\toprule
\textbf{Location} & \textbf{Count} & \textbf{Mean} \\
\midrule
Inside US & 6,808,489 & \textbf{0.75} $\pm$ 0.43 \\
Outside US & 11,050,903 & 0.74 $\pm$ 0.44 \\
\bottomrule
\end{tabular}
}
\label{tab:location_stats}
\end{table}

\begin{table}[ht]
\centering
\caption{Model performance by quiz location. RF is the random forst model with user and species context (\textit{RF} \texttt{U+S}), MLP is the MLP with user, species and ResNet image features (\textit{MLP} \texttt{U+S+Img ResNet}). Error bars are 2-sigma.}
\vspace{5pt}
\resizebox{0.8\columnwidth}{!}{%
\begin{tabular}{l l c c c c}
\toprule
\textbf{Location} & \textbf{Model} & \textbf{Binary - AP} & \textbf{Binary - Macro Acc} & \textbf{MC - Acc Full} & \textbf{MC - Acc Inc} \\
\midrule
Inside USA & RF & \textbf{0.86} & \textbf{85.34} & -- & -- \\
& & \textbf{$(\pm 0.00)$} & \textbf{$(\pm 0.01)$} & & \\
Outside USA & RF & 0.84 & 83.27 & -- & -- \\
& & $(\pm 0.00)$ & $(\pm 0.01)$ & & \\
\midrule
Inside USA & MLP & \textbf{0.79} & \textbf{84.30} & \textbf{76.71} & \textbf{25.90} \\
& & \textbf{$(\pm 0.02)$} & \textbf{$(\pm 0.59)$} & \textbf{$(\pm 0.33)$} & \textbf{$(\pm 1.94)$} \\
Outside USA & MLP & 0.71 & 81.92 & 73.41 & 23.98 \\
& & $(\pm 0.02)$ & $(\pm 0.34)$ & $(\pm 0.28)$ & $(\pm 1.23)$ \\
\bottomrule
\end{tabular}
}
\label{tab:combined_location_performance}
\end{table}

\subsection{Learning Dynamics}\label{app:learning_dynamics}

To gain insights into learning dynamics, we investigated model performance given different amounts of participant context. For this, we sorted the test set by context length and divided it into quintiles on a logarithmic scale. 
\cref{tab:quantile_performance} shows the result of our best performing binary model RF \texttt{U+S}. 
This reveals a clear monotonic improvement in both AP and macro accuracy as the amount of user context increases, suggesting that the model effectively integrates participant context into its predictions. Notably, the incremental gains decrease, with the sharpest rise observed between Q1 and Q2 and the smallest between Q4 and Q5, indicating diminishing returns of added context.

\begin{table}[ht]
\centering
\caption{Binary task performance of the random forest with user and species context (RF \texttt{U+S}) on the test set, stratified by context length quintiles on a logarithmic scale. Error bars are 2-sigma.}
\vspace{5pt}
\resizebox{0.8\columnwidth}{!}{%
\begin{tabular}{l c c c c c}
\toprule
\textbf{Metric} & \textbf{Q1} & \textbf{Q2} & \textbf{Q3} & \textbf{Q4} & \textbf{Q5} \\
\midrule
Binary AP & 0.62 & 0.77 & 0.88 & 0.93 & \textbf{0.96} \\
& ($\pm$0.0002) & ($\pm$0.0002) & ($\pm$0.0001) & ($\pm$0.0001) & ($\pm$0.0001) \\
\midrule
Binary Macro Accuracy & 67.25 & 80.09 & 87.73 & 91.73 & \textbf{93.56} \\
& ($\pm$0.0247) & ($\pm$0.0173) & ($\pm$0.0103) & ($\pm$0.0075) & ($\pm$0.0090) \\
\bottomrule
\end{tabular}
}
\label{tab:quantile_performance}
\end{table}

\subsection{Image Classification}\label{app:image_class}

To understand how informative our image features are, we evaluate multiple-choice species classification (\ie not participant guesses) directly from image features by training an one-layer MLP to predict the true species. 
The setup matches the image-only MLP (\textit{MLP} \texttt{Img}) runs, except that the model is not provided the ground-truth species and the target is the true species rather than the participant’s guess. 
Results are presented in \cref{tab:image_classification}. 
DINOv2 features achieve 88.8 ± 0.2\% accuracy, outperforming ResNet-50 by 13.9 percentage points (74.9 ± 0.4\%) with low variability across runs, indicating a strong advantage for the DINOv2 features.

\begin{table}[ht]
\centering
\caption{Species classification on the Multiple Choice task using an MLP with ResNet-50 or DINOv2 features. Error bars are 2-sigma over 10 training runs. }
\resizebox{0.6\columnwidth}{!}{%
\begin{tabular}{lcc}
\toprule
\textbf{Metric} & \textbf{ResNet-50} & \textbf{DINOv2} \\
\midrule
Multiple Choice Accuracy (\%) & 74.90 $\pm$ 0.40 & \textbf{88.80 $\pm$ 0.20} \\
\bottomrule
\end{tabular}
}
\label{tab:image_classification}
\end{table}

\section{Additional Dataset Details}

\begin{figure}[t]
\centering
\includegraphics[width=0.5\textwidth]{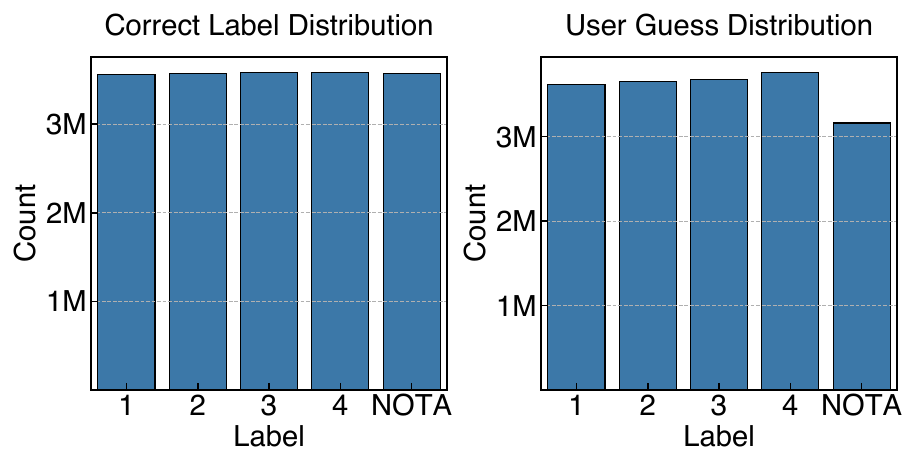}
\caption{Distributions of true labels across quiz choices and participant responses.}
\label{fig:hist_user_labels}
\end{figure}

\begin{figure}[t]
\centering
\includegraphics[width=\textwidth]{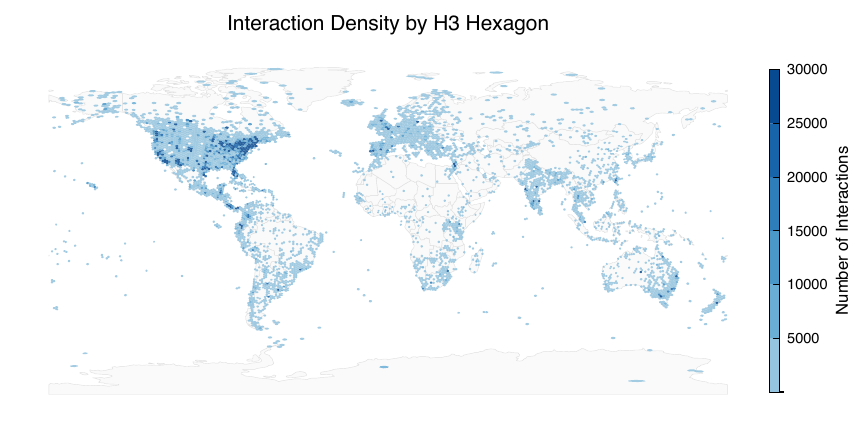}
\vspace{-20pt}
\caption{World map with Hex 3 polygonal bins representing quiz locations, where color intensity encodes the number of interactions per location cell.}
\label{fig:world_map}
\end{figure}

\begin{figure}[t]
\centering
\includegraphics[width=\textwidth]{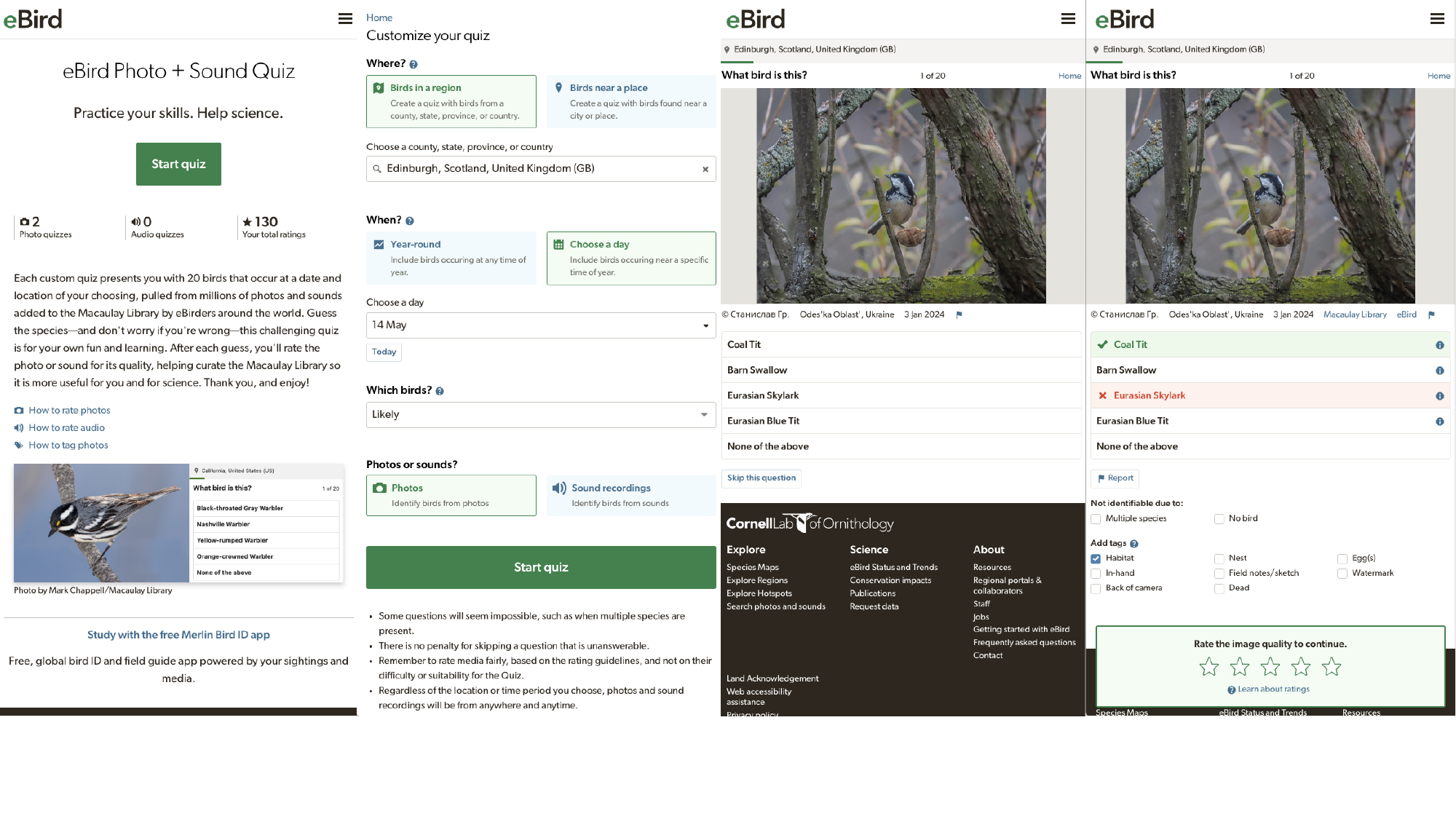} 
\vspace{-40pt}
\caption{Screenshots of the participant interface from the eBird quiz~\cite{eBirdQuiz} From left to right: (1) Quiz introduction page, presenting the task and linking to rating guidelines; (2) Customization page, where participants select quiz parameters such as location, time, species prevalence, and media type; (3) Question page, showing a bird image with five label choices and a skip option; (4) Feedback page, revealing the correct label and prompting a quality rating of the image.}
\label{fig:screenshots_feedback}
\vspace{-15pt}
\end{figure}

In \datasetname{}, the order of choices is random, as can be seen in~\cref{fig:hist_user_labels}. Participants' guesses, on the other hand, slightly disfavor the ``None of the above option'' option, and slightly favor the fourth option. 
Quiz location are chosen around the globe, as shown in~\cref{fig:world_map}, which shows quizzes' at H3 Hex 3 locations with interaction density. The distribution favors populated areas, and the global north.
\cref{fig:screenshots_feedback} shows four screenshots of the quiz website. First, the participant sees a general introduction to the quiz. Next, participants configure the quiz by selecting the place, time, and media type. Each of the 20 quiz questions presents a bird image alongside five answer choices; after responding, participants receive feedback and may rate the image quality.

{\bf Existing Knowledge Tracing Datasets.} 
In \cref{tab:kt_datasets} we summarize existing knowledge tracing datasets.  
We can see that our \datasetname dataset is significantly larger and more diverse in terms of the number of concepts contained withing compared to existing visual knowledge tracing datasets. 

\section{Additional Implementation Details}\label{appendix:implementation_details}

{\bf Input Encoding Transformer Models}
For learner \(l\), each historical interaction \( h^l_s \in \mathcal{H}^l_t \) is encoded as: 
\begin{equation}
h^l_s =  (\textsf{TOK}^C,\ y^l_s,\ \textsf{TOK}^O,\ \mathbf{c}^l_s,\ \textsf{TOK}^A,\ r^l_s), \quad \text{for } s \in [\max(0, t - W), t)],
\end{equation}
where $\textsf{TOK}^C$, $\textsf{TOK}^O$, and $\textsf{TOK}^A$ are defined as special tokens and are used to encode the type of information, specifically correct, options, and answers respectively, encoded in the next tokens. 
The model input consists of all interactions in the lookback window followed by the current question:
\begin{equation}
s^l_t = (h^l_{\max(0, t - W)},\ \textsf{TOK}^S,\ h^l_{\max(0, t - W)+1},\ \textsf{TOK}^S,\ \dots,\ h^l_{t-1},\ q^l_{t}),
\end{equation}
where $\textsf{TOK}^S$ indicates the separation token. Here, the current question \(q^l_{t}\) includes the correct species, the available choices and a special prompt token, but crucially not the actual participant answer: 
\begin{equation}
q^l_{t} = (\textsf{TOK}^C,\ y^l_{t},\ \textsf{TOK}^O,\ \mathbf{c}^l_{t},\ \textsf{TOK}^U), 
\end{equation}
where $\textsf{TOK}^U$ represents the participant's answer token type. 

In the classification model, each candidate answer \( c^l_{{t}, j} \) is appended to the question input, resulting in \(s^l_{t, j} = (q^l_{t},\ c^l_{{t}, j})\), for \(j = 1, \dots, K\). 
A classification token is prepended to the full sequence before encoding.

{\bf Training Configuration.} For the seq2seq model, we use Google's T5-Efficient-TINY-NL32 \cite{Tay2021Sep}, pretrained on the English C4 Corpus~\cite{dodge-etal-2021-documenting}. For the MCC approach, we use TinyBERT~\cite{Jiao2019Sep}, which is distilled from pretrained and fine-tuned BERT~\cite{Devlin2019Jun} teacher. \textit{LM Seq2seq} and \text{LM MCC} are consequently on a server equipped with 2 AMD EPYC 7763 64-Core Processor 1.8GHz (128 cores in total, 32 used for training), 1TiB RAM and 4 NVIDIA A100-SXM-80GB GPUs, with bf16 precision. 

We train each language model for 10 hours, using AdamW \cite{Loshchilov2017Nov} with a learning rate of \( 5 \times 10^{-5} \), weight decay of 0.01, maximum gradient norm of 1, and 4 gradient accumulation steps. Seq2seq models use a batch size of 512, while MCC models use 256. We evaluate every 200 steps on a validation subset of 300,000 examples, and select the best epoch based on validation performance. 

The MLP models (\textit{MLP} \texttt{Img}, \textit{MLP} \texttt{U+S+Img} and the image classification MLP models) are trained with hidden size 250, batch size 65,792, learning rate 0.001, and Adam \cite{Kingma2014Dec} for 3 epochs. They train on GPU in 2 hours with images and 15 minutes without images. Binary classifiers and confusion-prior baselines are trained using CPU in less than an hour in total. Results for runs with multiple seeds are shown in~\cref{tab:grouped_combined_metrics}.
Logistic regression, random forests and XGBoost~\cite{Chen:2016:XST:2939672.2939785} are trained using scikit-learn package~\cite{pedregosa2011scikit} default parameters. 
For all models receiving user context that are not the transformer models, the lookback window is set to the full history. For transformer models, is is set to 50 questions. 
The knowledge tracing baselines are trained using the pyKT package~\cite{Liu2022Jun} with default parameters. All baselines presented in \cref{tab:pykt_results_full} only predict binary outcomes, and receive user context.

{\bf Engineered features.} For our binary classifiers and the simple MLP, context is encoded through additional features. For user context, these are a user's average accuracy, user's average accuracy on the image species, log-transformed counts of how often the user has seen the species in the past, and how many questions the user has answered on the same location. 
Additionally, boolean indicators are provided for whether the user is in their geographic focus region by Hex 3 location, their spatio-temporal focus by Hex 3 location and whether the user has improved in the past in any location over a 20 question sliding window. 
Species context is provided as average accuracies for the species and the choices. For our language models, user context is provided in form of tokenized history sequences. The history lookback window is set to 50 for our language models, and to the full history for all other models.

\vspace{-5pt}
\section{Media Use}
\vspace{-5pt}

We used the following recordings from Cornell Lab of Ornithology | Macaulay Library: \cref{fig:quiz_examples} uses  
    ML614845753, ML624914011 and ML624836085. \cref{fig:dataset_quality_stats} upper row uses ML615927847, ML621578731, ML617550217 and ML621294128, lower row uses  ML39633601, ML50619491, ML38293181 and ML226495281. \cref{fig:confused_pairs} upper row uses  ML30091521,  ML117787821, ML302310521, ML83984151,  ML141517111 and ML284199291, lower row uses ML51777001, ML26854421, ML301728521, ML290513131, ML50787721 and ML174404171. \cref{fig:screenshots_feedback} uses ML463868861 and ML613090562.

\end{document}